\documentclass[10pt,twocolumn,letterpaper]{article}

\usepackage{cvpr}              % To produce the CAMERA-READY version
\definecolor{cvprblue}{rgb}{0.21,0.49,0.74}
\usepackage[pagebackref,breaklinks,colorlinks,allcolors=cvprblue]{hyperref}
\usepackage{multirow}
\usepackage{graphicx}

\def\paperID{7265} % *** Enter the Paper ID here
\def\confName{CVPR}
\def\confYear{2026}

\title{UniArt: Unified 3D Representation for Generating 3D Articulated Objects \\ with Open-Set Articulation}

\author{
  \textbf{Bu Jin}$^{1}$\quad \textbf{Weize Li}$^{2}$\quad \textbf{Songen Gu}$^{2}$\quad \textbf{Yupeng Zheng}$^{2}$\quad \textbf{Yuhang Zheng}$^{2}$\quad \textbf{Zhengyi Zhou}$^{2}$\quad \textbf{Yao Yao}$^{3}$\\
  $^{1}$Hong Kong University of Science and Technology \quad $^{2}$Tsinghua University\quad $^{3}$Nanjing University\\
  \texttt{bjinaa@connect.ust.hk}\quad \texttt{yaoynju@gmail.com}
  }

\begin{document}
\maketitle

% \twocolumn[{%
% \renewcommand\twocolumn[1][]{#1}%
%     \begin{center}
%             \vspace{-0.7cm}
%             \captionsetup{type=figure}
%             \includegraphics[width=0.85 \textwidth]{fig/iclr_teaser.png}
%             \vspace{-0.1cm}
%             \caption{\small We propose UniArt, the first non-retrieval, diffusion-based framework that generates robot-ready articulated 3D objects from a single image, enabling open-set generalization for scalable simulation and manipulation.}
%             % \label{fig:teaser}
%     \end{center}
% }]

\begin{abstract}
Articulated 3D objects play a vital role in realistic simulation and embodied robotics, yet manually constructing such assets remains costly and difficult to scale. 
In this paper, we present \textbf{UniArt}, a diffusion-based framework that directly synthesizes fully articulated 3D objects from a single image in an end-to-end manner. Unlike prior multi-stage techniques, UniArt establishes a unified latent representation that jointly encodes geometry, texture, part segmentation, and kinematic parameters. We introduce a reversible joint-to-voxel embedding, which spatially aligns articulation features with volumetric geometry, enabling the model to learn coherent motion behaviors alongside structural formation. Furthermore, we formulate articulation type prediction as an open-set problem, removing the need for fixed joint semantics and allowing generalization to novel joint categories and unseen object types.
Experiments on the PartNet-Mobility benchmark demonstrate that UniArt achieves state-of-the-art mesh quality and articulation accuracy. 
\end{abstract}    
\section{Introduction}
\label{sec:intro}

\begin{figure}[!t]
    \centering
    \includegraphics[width= \linewidth]{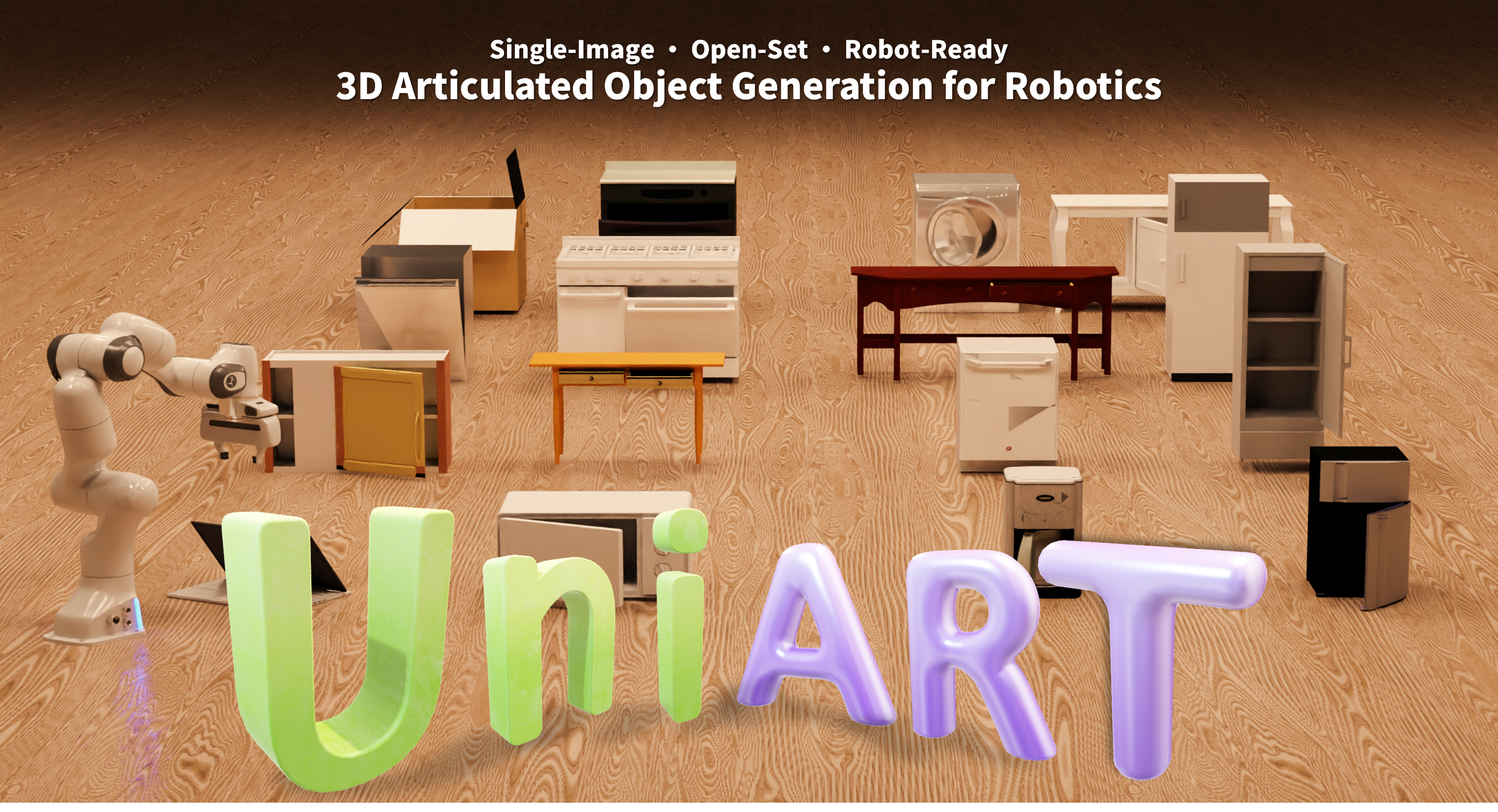}
    \caption{\small We propose UniArt, a novel diffusion-based framework that generates robot-ready articulated 3D objects from a single image, enabling open-set generalization for scalable simulation and manipulation.}
    \label{fig: overview}
    \vspace{-0.3cm}
\end{figure}

3D articulated objects \cite{quigley2015programming} are core components of mechanical systems, ranging from common doors in daily life to complex joint mechanisms in robotic grippers. 
Unlike static rigid 3D assets, articulated objects exhibit inherent part-level structures and motion patterns, enabling dynamic interactions such as opening a drawer, swiveling a chair, or operating scissors. 
Precise modeling of these structures ~\citep{tseng2022cla, liu2023paris, weng2024neural, iliash2024s2o, mandi2024real2code, liu2024singapo, liu2024cage, wu2025dipo} not only supports the development of high-fidelity simulation environments ~\citep{chen2024rapid, li2023synthesizing, li2024advances, luo20233d} but also paves the way for accurate dynamic analysis in embodied robotics~\citep{yang2024physcene,yang2024holodeck,geng2025roboverse}.
Considering that acquiring detailed annotations for such objects remains highly labor-intensive and struggles to keep pace with the growing diversity of objects required by embodied robots, an automated generation solution for articulated objects is needed.

Traditional methods for articulated object generation~\citep{liu2024singapo, liu2024cage, wu2025dipo,deng2024articulate} follow a three-stage pipeline. First, articulation parameters, including part bounding boxes and semantic labels, are predicted from input images.
Next, a corresponding part geometry is retrieved from a pre-existing asset library.
Finally, the retrieved parts are assembled into a complete object. 
While retrieval-based methods~\cite{jiang2022ditto,gao2025meshart,su2025artformer,qiu2025articulate, liu2025artgs, shen2025gaussianart} provide a shortcut for generating new articulated models, they introduce several critical limitations: geometric misalignment due to imperfect part matching, limited diversity bounded by the pre-defined asset collection, and poor generalization to objects outside the training distributions.
These issues hinder the deployment of such methods in open-world scenarios, where objects exhibit vast variations in form, function, and material.

% challenges for generative models for articulated objects generation 
Recent progress in 3D generation \cite{xiang2025structured, wu2024direct3d, wu2025direct3d} has shown that diffusion and transformer-based frameworks can synthesize novel 3D objects from inputs such as text or images. Some works like ArtFormer \cite{su2025artformer} have demonstrated that conditional generation of articulated objects is feasible, which effectively addresses the geometric misalignment caused by retrieval. However, these approaches usually utilize a part‑wise generation strategy, which generates each sub‑part individually and then assembles these parts into a complete mesh. This compositional process often leads to global structural inconsistency across parts, such as discontinuities at part boundaries, misaligned joint interfaces, and overall incoherent geometry.

\begin{figure*}[t]
\centering
\includegraphics[width=0.9\textwidth]{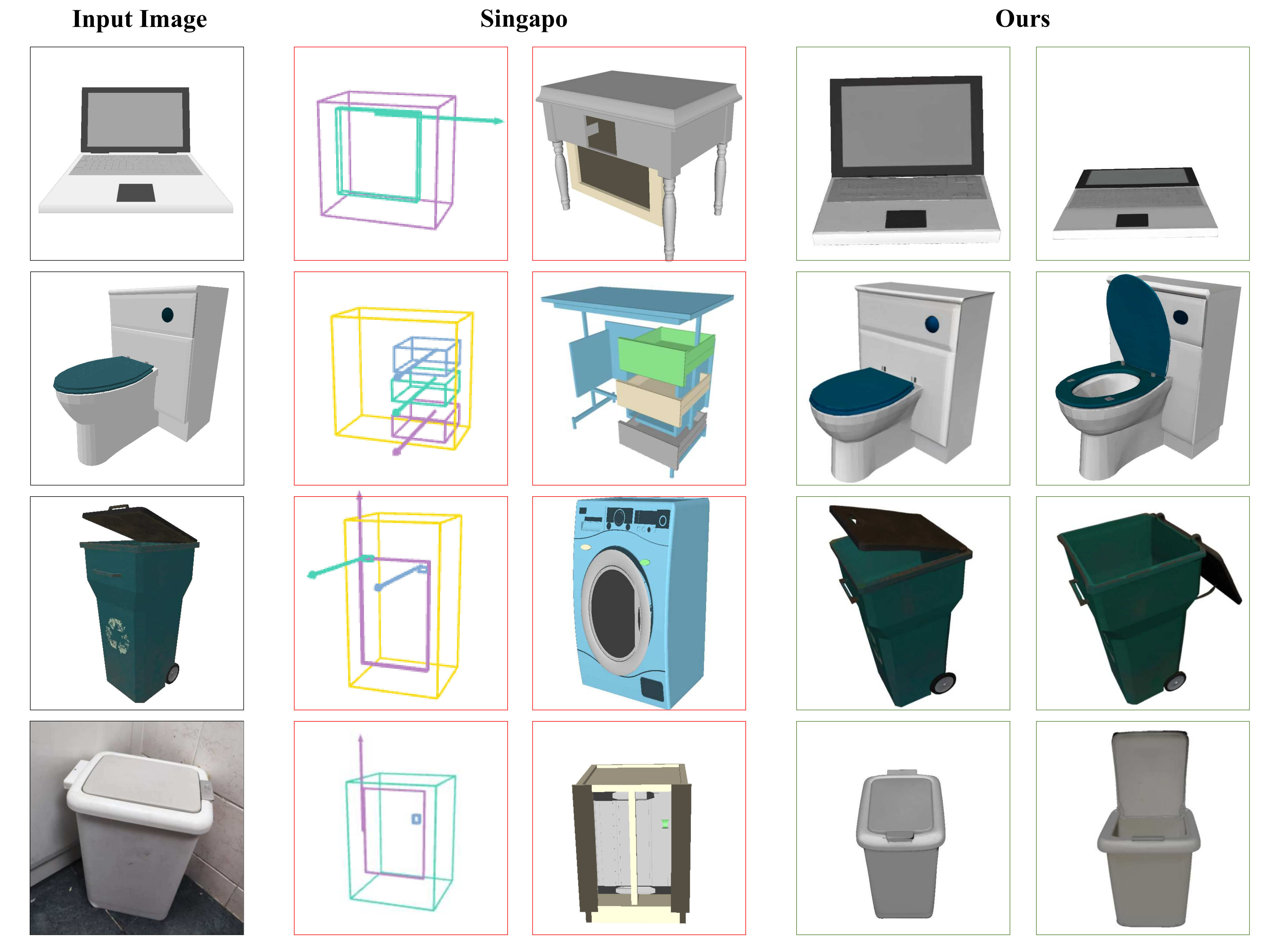}
\caption{Qualitative results on unseen categories. It can be observed that the articulated objects generated by our method exhibit good consistency with the input images in both appearance and geometry, while previous retrieval-based methods fail to generate sound results.}
\label{fig:openset_qualitative}
\vspace{-3mm}
\end{figure*}

Moreover, another critical challenge for previous methods is that they rely heavily on predefined semantic joint categories. The parts can only be generated or retrieved based on fixed semantic labels, which restricts the diversity of the generated objects. For example, the semantic labels in Singapo \cite{liu2024singapo} include ``base, door, drawer, handle, knob, tray'', meaning that the generated articulated objects can only contain several fixed categories. Although ArtFormer \cite{su2025artformer} utilizes the text embedding instead of class embedding to increase their performance on unseen object categories, the fixed joint categories prevent their application on more diverse objects with totally different joint categories (such as the toilet in Fig.~\ref{fig:openset_qualitative}).

To address these crucial issues, we propose UniArt, an end-to-end framework that synthesizes articulated objects directly without relying on part retrieval.
Our paradigm shift centers on rethinking two core concepts.
% correspond to the former paragraphs
% 1. compared to retrieval-based methods
\textbf{First}, we propose a generative model that directly generates the whole articulated objects instead of generating or retrieving each part independently. 
% 2. compared to multi-stage generative methods
\textbf{Second}, we introduce a reversible joint-to-voxel embedding method that spatially grounds kinematic parameters with geometry. Along with the previous appearance-to-voxel approach \cite{xiang2025structured}, UniArt is able to create a unified latent representation, named UniArt latents, which jointly encodes the geometric information, visual appearance, part segmentation, and articulation parameters.
% 3. compared to methods with fixed-category joint
\textbf{Third}, we treat articulation type prediction as an open-set problem, eliminating the need for predefined joint semantic labels. Specifically, while previous methods \cite{gao2025meshart, su2025artformer} utilize semantic labels when generating sub-parts, such explicit high-level semantics are not necessary during the generative process. Instead, our model directly utilizes learned, part-based features to infer articulation types, enabling generalization to novel objects and joint categories.

Comprehensive evaluations on the PartNet-Mobility benchmark demonstrate that UniArt outperforms existing baselines significantly in terms of mesh quality and articulation accuracy, particularly under open-set conditions.

Our contributions can be summarized as follows:
\begin{itemize}
% a unified conditional generation framework
% texture?
\item We propose UniArt, a novel diffusion-based model that unifies geometry synthesis, texture generation, part segmentation, and articulation prediction within a single latent representation, which directly generates the whole articulated objects instead of generating or retrieving each part independently.
\item We introduce a reversible joint-to-voxel embedding that spatially grounds kinematic parameters with geometry, and treat joint type prediction as an open-set problem without predefined labels.
\item We develop a voxel-level latent space that simultaneously encodes geometry occupancy, visual appearance, part-aware features and articulation constraints, ensuring mutual consistency during generation.
% \item We introduce UniArt latent representations that jointly encode object geometry, appearance, part segmentation, and articulation parameters within a diffusion-based architecture.
% \item We propose a diffusion-based framework, UniArt, which models the UniArt latent with a Physaware Attention Module that enables joint generation of structural and open-set kinematic information of articulated objects.
\item We show through comprehensive experiments that our method substantially advances the state of the art in articulated object generation.
\end{itemize}
\section{Related Works}
\label{sec:related}

\subsection{Reconstruction-based 3D Articulated Object Creation.}
The reconstruction methods ~\citep{tseng2022cla, liu2023paris, weng2024neural, iliash2024s2o, mandi2024real2code,kim2025screwsplat} typically rely on multi-view or multi-state inputs to recover part-level geometry and articulation parameters. On the basis of NeRF, CLA-NeRF~\citep{tseng2022cla} utilizes a component segmentation field to predict the categories of each component of the articulated object, in order to perform view synthesis, component segmentation, and joint pose estimation of unknown articulated poses. PARIS~\citep{liu2023paris} presents a self-supervised architecture for part-level reconstruction and motion analysis of articulated objects, achieving significant improvements in shape reconstruction and motion estimation without requiring 3D supervision. Real2Code~\citep{mandi2024real2code} utilizes the knowledge of LLM to get the articulation parameters, also requiring multi-view images as input.
ArtGS~\citep{liu2025artgs} employs a strategy from coarse to fine, using the Hungarian algorithm to match Gaussian spheres in different states and cleverly establish corresponding relationships between different states of objects. GaussianArt~\citep{shen2025gaussianart} instead introduces a unified representation based on articulated 3D Gaussian primitives, generating good reconstruction results with correct articulation parameters. Although reconstruction-based methods provide good results for generating new articulated objects, they rely on multi-view or multi-state inputs, which are not easily accessible for scalable data construction.
In contrast, our method requires only a single image as input. This substantially reduces input complexity, enabling scalable data collection and facilitating large-scale robot training.

\subsection{Generative 3D Articulated Object Creation.}
Recent progress in 3D generation~\citep{li2025sparc3d, chen2025dora, li2024craftsman3d, ren2024xcube, tochilkin2024triposr, wang2023rodin, zhao2025hunyuan3d, wu2024direct3d, wu2025direct3d} has enabled applications in 3D Articulated Object Creation.
The generative 3D articulated object creation aims to generate part-level geometry and articulation parameters through a single image. Previous generative articulated object creation methods typically rely on retrieval, where a corresponding part geometry is retrieved from a pre-existing asset library.
Articulate-Anything~\citep{le2024articulate} first converts static 3D assets into articulation-ready models by retrieving the most similar asset from the library and generates articulation parameters through reinforcement learning.
URDFormer~\cite{chen2024urdformer} attempts to directly infer an interactive URDF from the images. Likewise, OPDMulti~\citep{sun2024opdmulti} localizes movable parts and estimates motion parameters from a single image.
~\cite{le2024articulate, liu2024singapo, liu2024cage, wu2025dipo} generate articulated object structures from inputs such as images and graphs. However, these methods often depend on mesh retrieval from a fixed database, which restricts both the variety of generated objects and the adaptability to subjective user specifications. Artformer \cite{su2025artformer} instead proposes a transformer-based generative model to produce novel geometries for sub-parts to extend the geometric diversity of the generated objects, while it still suffers generating objects beyond the predefined labels. In contrast, our approach aims to synthesize new objects .
\section{Problem Formulation}
\label{formu}

\begin{figure*}[t]
    \centering
    \includegraphics[width=1.0\linewidth]{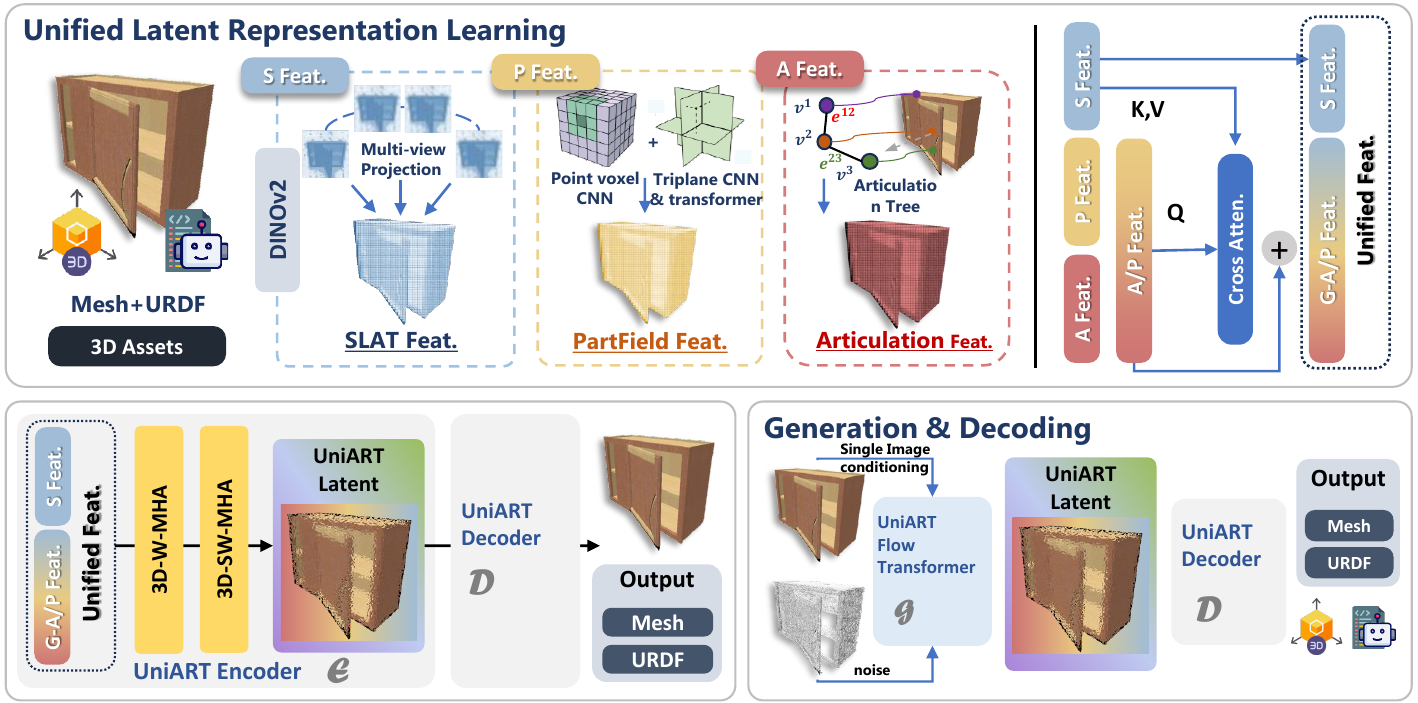}
    \caption{\small Overview of \textbf{UniArt}. We reformulate the articulated object creation task and introduce UniArt latent representations that jointly encode object geometry, appearance, part segmentation, and articulation parameters within a diffusion-based architecture.}
    \label{fig: overview}
    \vspace{-0.3cm}
\end{figure*}

Existing retrieval-based pipelines for articulated object generation first predict part proposals, then search a finite repository of pre-rigged assets for the closest matches, and finally stitch the retrieved parts together. Due to the fact that each component is copied verbatim from the database, these methods cannot guarantee geometric continuity at part boundaries, inherit whatever material and joint types the database provides, and fail gracefully when the target object falls outside the predefined taxonomy. In contrast, we reformulate the task as a fully generative problem, synthesizing geometry, part structure, and articulation parameters in a continuous latent space so that every piece is produced with mutual consistency and the design space is no longer bounded by the content of a repository.

Formally, we denote the target articulated asset as:
\begin{equation}
    A = (M, S, U)
\end{equation}
where $M$ is a watertight triangle mesh, $S$ is a part segment mask, and $U$ is a URDF specification~\citep{quigley2015programming} containing joint type $T$, connection topology $J$, axis direction $A$, joint limits $L$ and body assignments $B$ of each part:
\begin{equation}
    U = (T, J, A, L, B)
\end{equation}
The input is a single RGB image $I$.

Retrieval-based methods first infer an articulation tree from the given image $I$, where each part contains the bounding boxes $B=\{b,c\}$ that list each part’s 3D bounding box $b\in\mathbb{R}^6$ and semantic label $c$. They then select parts from a part repository $\mathcal{P}=\{(M_i,\ell_i)\}_{i=1}^{N}$ that provides mesh geometry $M_i$, and the corresponding label $\ell_i$, the method filters the database by label, $\mathcal{D}_{c}=\{i\mid\ell_i=c\}$, and retrieves the closest candidate by box similarity,
\begin{equation}
\hat{M}=\arg\min_{i\in\mathcal{D}_{c}} d_{\text{size}}\bigl(\operatorname{box}(M_i),\,b\bigr).
\end{equation}
The selected parts are rigidly aligned to the predicted boxes and concatenated to yield
\begin{equation}
A^{\text{retr}}=\mathcal{A}\bigl(\{\hat{M}_k\}_{k=1}^{K},B\bigr)=(M^{\text{retr}},S^{\text{retr}},U^{\text{retr}}).
\end{equation}
where $K$ is the number of parts in the articulation tree.
We can see that every component of $A^{\text{retr}}$ is copied from $\mathcal{D}$, and thus the output can never exceed the geometric fidelity, material diversity, or articulation vocabulary encoded in the repository, and inevitable misalignments across part boundaries produce visual and kinematic inconsistencies.

Instead, in the generative method, we learn a conditional diffusion model over a single latent vector $z\in\mathbb{R}^{d}$, dubbed UniLatent, that jointly encodes geometry, appearance, part structure, and the articulation tree. The forward process corrupts $z$ with Gaussian noise, while the reverse process produces $z_0\sim p_\theta(z\mid I)$. A shared decoder then deterministically maps $z_0$ into an articulated asset $A=(M,S,U)$ through three parallel heads$G_{\text{geo}}$, $G_{\text{seg}}$, $G_{\text{art}}$, formulated as:
\begin{equation}
(M, S, U)=\bigl(G_{\text{geo}}(z_0),\;G_{\text{seg}}(z_0),\;G_{\text{art}}(z_0)\bigr),
\end{equation}
yielding the overall distribution
\begin{equation}
p_\theta(A\mid I)=\int \delta\!\bigl(A-G(z)\bigr)\,p_\theta(z\mid I)\,dz,
\end{equation}
where $G=(G_{\text{geo}},G_{\text{seg}},G_{\text{art}})$ and $\delta(\cdot)$ is the Dirac delta. In this setting, $p_\theta$ is learned in a continuous latent space where UniArt can produce infinitely many geometries and articulation patterns that are not restricted to the discrete set $\mathcal{D}$, while its open-set formulation removes the need for fixed semantic labels during training and enables robust performance on previously unseen categories.
\section{Methods}
\label{method}
As illustrated in Fig.~\ref{fig: overview}, our goal is to generate articulated objects in a unified framework that simultaneously produces geometry meshes, part-level segmentation, and articulation parameters. To support this generation process, we should parameterize the geometry, part segment, and articulated structure into vectors that can be the target of the diffusion. We introduce how we parameterize the articulated objects into a latent space in Sec .~\ref {sec: parameterization}. Then we introduce our variational autoencoder that compresses the parameterized articulated object into the latent space in Sec .~\ref {sec: vae}. Finally, we illustrate the generation process in Sec .~\ref {sec: generation}. 

\subsection{Articulated Object Parameterization}  
\label{sec: parameterization}
As mentioned before, we adopt the URDF representation for parameterization of articulated objects, which represents each object as a connected graph in which nodes denote links (parts) and edges denote joints. We follow the common practice of NAP~\citep{lei2023nap} and assume the kinematic graph is connected with no cycles and each edge is a screw joint with at most one prismatic translation and one revolute rotation, covering most real-world articulated objects~\citep{xiang2020sapien, wang2019shape2motion}.

A central challenge in parameterizing articulated objects is how to encode the joint-level kinematics in a representation that is spatially compatible with voxelized geometry. 
While URDF specifies each part node and its connecting joints in symbolic form, the geometry mesh is voxelized into a dense feature grid. These attributes must be grounded into a continuous volumetric tensor for unified encoding. We address this via a joint-to-voxel embedding scheme.

We describe the URDF parameters $U=\{u^0, u^1,...,u^{K-1}\}$ as a graph with $K$ nodes and each $u^i$ attributes encoding joint type $t^i$, axis $a^i\in\mathbb{R}^6$ and motion limits $l^i=(l_{\min},l_{\max})$, denoted as: $u^i=(t^i, a^i, l^i)$.
Unlike traditional retrieval-based approaches that often rely on predefined semantic labels of links (like base, door, drawer, handle, knob, tray in SINGAPO~\citep{liu2024singapo}), we intentionally exclude such categorical annotations in our formulation. This design choice avoids introducing bias toward a fixed set of link categories and instead encourages the model to generate links and kinematic structures that are not limited to predetermined templates, thereby improving generalization to novel articulation morphologies.
We serialize $U$ into a sparse adjacency tensor $c\in\mathbb{R}^{9}$, which serves as the articulation representation of a joint. 
For the connection graph $J$, we form the adjacency tensor $J \in \{0,1\}^{K\times K}$.  

Then we conduct joint-to-voxel projection, which aligns the joint parameters with the corresponding mesh and part structure. For each node $i \in \{0, \dots, K-1\}$, we associate its attributes to the edge that connects this node to its parent in the kinematic tree. This design ensures that the parameters are naturally interpreted as governing the motion of the child link with respect to its parent link. 

The attributes carried by the edge are then projected onto the 3D voxel space that represents the geometric occupancy of the child part. In this process, all active voxels belonging to the mesh region of the child part inherit the same parameter assignment, thereby embedding the kinematic constraints directly into the spatial representation of the part. This provides a unified voxel-level representation where both geometric and kinematic information co-exist, enabling subsequent models to jointly reason about structure and motion. 
Since each node in the articulation tree has exactly one parent, this assignment is reversible during decoding. Given voxel-level encoding, we can uniquely recover the corresponding node attributes and rebuild the parent-child relationships. This property is essential to guarantee consistency between the learned voxelized representation and the original kinematic graph structure.

\begin{table*}[t]
\centering
\caption{\small Comparison of generation quality and graph prediction accuracy on \textbf{PartNet-Mobility} test set. ``-'' represents that the code is not available at present. ``RS'' and ``AS'' represent resting states and articulated states, respectively.}
% \vspace{-0.2cm}
\resizebox{0.8\textwidth}{!}{ % 控制表格缩放
\begin{tabular}{l|cccccc}
\toprule
\multirow{2}{*}{Method} & \multicolumn{2}{c}{Appearance-Articulation} & \multicolumn{2}{c}{Gemoetry-Articulation} & \multicolumn{2}{c}{Shape-Articulation}  \\
\cmidrule(lr){2-3} \cmidrule(lr){4-5} \cmidrule(lr){6-7} 
& RS-$d_{\text{PSNR}}$ $\uparrow$ & AS-$d_{\text{PSNR}}$ $\uparrow$ 
& RS-$d_{\text{CD}}$ $\downarrow$ & AS-$d_{\text{CD}}$ $\downarrow$
& RS-$d_{\text{OpenShape}}$ $\uparrow$ & AS-$d_{\text{OpenShape}}$ $\uparrow$ 
\\
\midrule
URDFormer~\cite{chen2024urdformer} &12.31 & 10.45 & 0.4417 & 0.6910 & 0.0431 & 0.0374\\
NAP-ICA~\cite{lei2023nap} &14.27 & 12.74 & 0.0209 & 0.3473 & 0.0932 & 0.0872 \\
SINGAPO~\cite{liu2024singapo} &17.16 & 13.90 & 0.0191 & 0.1270 & 0.1073 & 0.0915 \\
DIPO~\cite{shen2025gaussianart}  &-  & - & 0.0132 & 0.0423 & - & -  \\
\midrule
\textbf{UniArt (Ours)}  & \textbf{28.52} & \textbf{23.77} & \textbf{0.0095} & \textbf{0.0376} & \textbf{0.1457} & \textbf{0.1176} \\
\bottomrule
\end{tabular}
}
\label{tab:main}
\vspace{-0.3cm}
\end{table*}

\subsection{UniArt VAE with Geometry-Articulation Interaction}
\label{sec: vae}
After obtaining the voxelized representation of both geometry and articulation, our next step is to learn a compact latent space that jointly captures structural and kinematic information. We construct a unified structured latent representation, named UniArt Latent, and utilize a variational autoencoder (VAE) tailored for this unified representation.

For each 3D asset, we first convert the mesh into a binary occupancy grid, resulting in a voxelized geometric feature $V_{geo}$ enriched with visual features by multiview average, following~\cite{xiang2025structured}. In parallel, the articulation representation introduced in the previous subsection is also voxelized, producing per-voxel articulation attributes. For the part representation, we utilize a pretrained model partfield \cite{liu2025partfield} to generate part-aware representations. The part representations and articulation attributes are added and voxelized, resulting in final articulation features $V_{art}$. All features are defined on the same voxel space with size $N$, where $N$ represents the total number of active voxels.

Instead of relying on naive concatenation, we introduce \textbf{part-wise UniArt attention} to dynamically align $V_{geo}$ and $V_{art}$. Specifically, we treat the articulation feature as the query and the geometric feature as key–value pairs:  
\begin{equation}
    F_{art} = \text{Attention}(Q = V_{art}, K = V_{geo}, V = V_{geo}) + V_{art},
\end{equation}
where the cross-attention modules aggregate motion-aware features that are consistent with the underlying geometric structure. The fused representation is enhanced with a residual connection to preserve original geometric detail.
These two feature types are channel-wise concatenated into a unified voxel feature map:
\begin{equation}
V = \text{Concat}(V_{geo}, F_{art})
\end{equation}
where each voxel is enriched with both spatial occupancy and articulation-aware information.
This design ensures that articulation information is selectively integrated depending on local geometry, encouraging the model to learn physically plausible correlations between part shape and its kinematic behavior.  

The unified feature $V \in \mathbb{R}^{N \times C}$, with $N$ active voxels and $C$ channels, is then passed through the VAE encoder $\mathcal{E}_{vae}$. The encoder employs attention layers to learn hierarchical spatial features while preserving the alignment between geometry and motion constraints. 
The sampled latent embedding $z$ is passed into the VAE decoder $\mathcal{D}_{vae}$, which reconstructs both geometry and articulation features simultaneously:
\begin{equation}
    \hat{V}_{geo}, \hat{V}_{art} = \mathcal{D}_{vae}(z).
\end{equation}
Unlike conventional VAE frameworks~\citep{cao2025physx} that separately encode physical or appearance properties, our decoder is optimized to jointly restore the voxelized structure and articulation. This design ensures that the model learns a latent space where geometry and motion are inherently entangled, facilitating more faithful morphology reconstruction.  

The VAE is optimized with a compound loss function:  
\begin{equation}
\mathcal{L}_{vae} = \mathcal{L}_{geo} + \mathcal{L}_{art} + \mathcal{L}_{kl}.
\end{equation}  
where $\mathcal{L}_{geo}$ measures the reconstruction fidelity of voxel occupancy, $\mathcal{L}_{art}$ supervises the recovery of articulation attributes, and $\mathcal{L}_{kl}$ is the Kullback–Leibler regularization term. Together, these terms encourage the VAE to disentangle structural and kinematic variations while maintaining a compact latent space suitable for downstream generation and inference tasks.

\subsection{Articulated Latent Generation}
\label{sec: generation}
After obtaining the fused latent representation from the VAE encoder, we aim to generate novel articulated objects with consistent geometry and articulation. We design a latent diffusion model that simultaneously models structural layout and articulation parameters. The generator is implemented as a rectified flow model, similar to~\cite{xiang2025structured}, and the training objective is the conditional flow matching objective:
$\mathcal{L}=\mathbb{E}_{t,x_0,\epsilon}||f(x,t)-(\epsilon-x_0)||^2_2$
where $f(x, t)$  is the conditional flow field that transports noisy samples to the clean latent distribution, $x_0$ is a latent from the VAE encoder, $\epsilon$ is Gaussian noise, and $t$ is the timestep.

% \begin{figure}[t]
% \centering
% \includegraphics[width=0.9\textwidth]{fig/qualitative.pdf}
% \caption{Qualitative results of UniArt. Since retrieval-based methods lack appearance information, we randomly applied different colors to distinguish each link. Our method exhibits better consistency in both appearance and geometry, while the results of SINGAPO~\citep{liu2024singapo} suffer from articulation error (\textcolor{red}{Red Box}), geometry inconsistency, and appearance inconsistency. }
% \label{fig:base_qualitative}
% \vspace{-0.3cm}
% \end{figure}
\section{Experiments}

\begin{figure*}[t]
\centering
\includegraphics[width=0.9\textwidth]{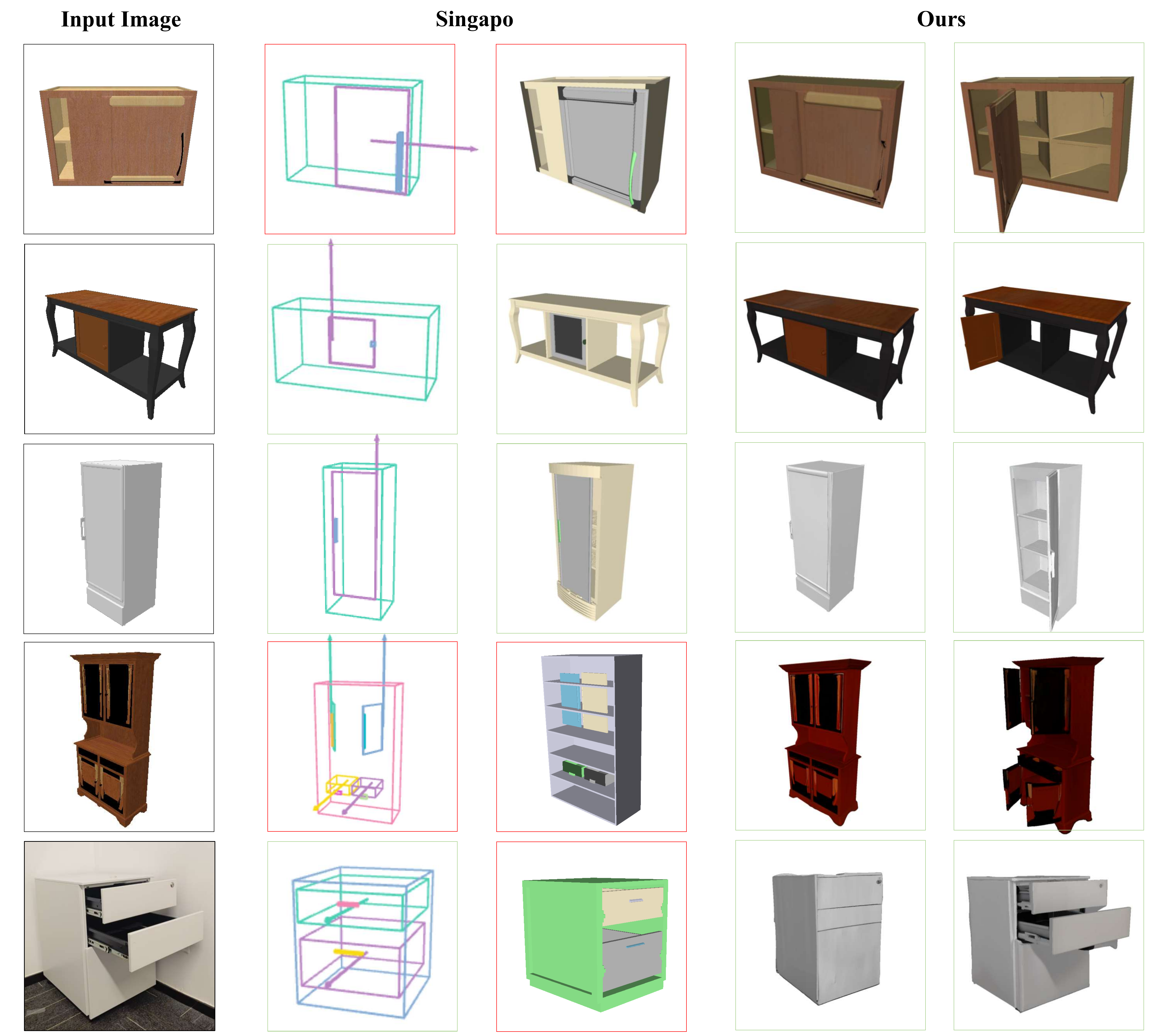}
\caption{Qualitative results of UniArt. Since retrieval-based methods lack appearance information, we randomly applied different colors to distinguish each link. Our method exhibits better consistency in both appearance and geometry, while the results of Singapo~\citep{liu2024singapo} suffer from articulation error (\textcolor{red}{Red Box}), geometry inconsistency, and appearance inconsistency. }
\label{fig:base_qualitative}
\vspace{-0.3cm}
\end{figure*}

We evaluate UniArt on the PartNet-Mobility benchmark, which provides a diverse set of articulated objects with ground-truth meshes, part annotations, and URDF parameters. 
Besides the common evaluation practice, we also conduct open-set evaluation. We split the dataset into seen categories (Storage, Table, Refrigerator, Dishwasher, Oven, Washer, and Microwave) and unseen categories (Bottles, Toilet, Chair, etc.) to test open-set generalization.

\subsection{Experimental Setup}
We follow the dataset split utilized in common evaluation practice~\citep{wu2025dipo, liu2024singapo}. 
During training, we augment datasets with random perturbations in part geometry, synthetic articulation parameter sampling within physically valid ranges. The total training samples are 45k. We utilize the AdamW optimizer with a learning rate of $1e-4$. Models are trained on 8 NVIDIA A100 GPUs with a batch size of 64. To ensure easier convergence, we initialize our model with the 3D geometric and visual prior from Trellis~\citep{xiang2025structured}. 

\subsection{Evaluation Metrics}
Previous works on articulated object generation primarily evaluate articulation accuracy. Most benchmarks assume that the geometry of retrieved parts is correct, and therefore ignore two essential aspects: visual fidelity and shape-image consistency, which affect perceptual quality in graphics and simulation.

As a result, existing evaluation protocols underestimate the challenges faced by generative models that must directly synthesize geometry, appearance, and kinematics. To provide a fair and comprehensive benchmark for generation-based methods, we introduce novel evaluation metrics based on the 3D generation task. We follow the common practice of 3D generative models and utilize PSNR, Chamfer Distance, and OpenShape~\citep{liu2023openshape} metrics to respectively measure the appearance, geometry, and shape-image alignment between generated meshes and conditional input images.
All metrics are computed over both resting states and articulated states, denoted as (RS-) and (AS-). For articulated states, we uniformly sample from the resting to the end state and compute the average metrics, following~\cite{liu2024singapo}.

\subsection{Main Results}
We report quantitative comparisons on the PartNet-Mobility test set in Table~\ref{tab:main}. The results demonstrate that our model, UniArt, consistently outperforms prior methods across all metrics, validating the effectiveness of our unified voxel–articulation representation and diffusion-based generation.
It is important to note that retrieval-based methods produce uncolored meshes. Thus, we assign ground-truth materials to the uncolored meshes, ensuring a consistent comparison across all models. 

In the resting state (RS-), UniArt achieves a PSNR of 28.52, improving over the SINGAPO~\citep{liu2024singapo} by 11.36, reflecting highly faithful texture reconstruction. With a Chamfer Distance of 0.0095, our method surpasses DIPO~\citep{shen2025gaussianart}, highlighting superior fidelity in static shape generation. UniArt also obtains the highest OpenShape score, showing better alignment between generated shapes and conditional images.

Across articulated states (AS-), performance gains remain substantial, where UniArt reaches 23.77, 9.87 higher than SINGAPO. This demonstrates robustness in preserving appearance even under large part motions. The Chamfer Distance and OpenShape similarity also outperform previous works, setting new state-of-the-art on articulated object generation. 

UniArt consistently achieves the best results in terms of appearance fidelity, geometric accuracy, and perceptual alignment. The gains in articulated states are particularly notable, showing that our unified voxel–articulation latent representation ensures stable geometry and motion consistency throughout the articulation process. The qualitative results are shown in Fig.~\ref{fig:base_qualitative}.

% \begin{figure*}[!ht]
% \centering
% \includegraphics[width=0.8\textwidth]{fig/real_robot.pdf}
% \caption{Application in the robotic manipulation.}
% \label{fig:real}
% % \vspace{-0.5cm}
% \end{figure*}

\begin{table*}[t]
\centering
\caption{Ablative results of generation quality and articulation prediction on Partnet-Mobility dataset.}
\resizebox{0.8\textwidth}{!}{ % 控制表格缩放
\begin{tabular}{cc|cccccc}
\toprule
\multicolumn{2}{c|}{Settings} & \multicolumn{2}{c}{Appearance-Articulation} & \multicolumn{2}{c}{Gemoetry-Articulation} & \multicolumn{2}{c}{Shape-Articulation}  \\
\cmidrule(lr){1-2} \cmidrule(lr){3-4} \cmidrule(lr){5-6} \cmidrule(lr){7-8}
 Uni-encoding & 3D Prior
& RS-$d_{\text{PSNR}}$ $\uparrow$ & AS-$d_{\text{PSNR}}$ $\uparrow$ 
& RS-$d_{\text{CD}}$ $\downarrow$ & AS-$d_{\text{CD}}$ $\downarrow$
& RS-$d_{\text{OpenShape}}$ $\uparrow$ & AS-$d_{\text{OpenShape}}$ $\uparrow$  \\
\midrule
$\checkmark$  &              & 13.24 & 11.76 & 0.0572 & 0.0763 & 0.0504 & 0.0427 \\ 
              & $\checkmark$ & 23.75 & 21.37 & 0.0149 & 0.0462 & 0.1175 & 0.1043 \\ \midrule
$\checkmark$  & $\checkmark$ & \textbf{28.52} & \textbf{23.77} & \textbf{0.0095} & \textbf{0.0376} & \textbf{0.1457} & \textbf{0.1176} \\
\bottomrule
\end{tabular}
}
\label{tab:ablation}
\vspace{-0.2cm}
\end{table*}

\begin{table*}[t]
\centering
\caption{Impact of different aggregation methods.}
\resizebox{0.8\textwidth}{!}{ % 控制表格缩放
\begin{tabular}{c|cccccc}
\toprule
\multirow{2}{*}{Aggregation} & \multicolumn{2}{c}{Appearance-Articulation} & \multicolumn{2}{c}{Gemoetry-Articulation} & \multicolumn{2}{c}{Shape-Articulation}  \\
\cmidrule(lr){2-3} \cmidrule(lr){4-5} \cmidrule(lr){6-7}

& RS-$d_{\text{PSNR}}$ $\uparrow$ & AS-$d_{\text{PSNR}}$ $\uparrow$ 
& RS-$d_{\text{CD}}$ $\downarrow$ & AS-$d_{\text{CD}}$ $\downarrow$
& RS-$d_{\text{OpenShape}}$ $\uparrow$ & AS-$d_{\text{OpenShape}}$ $\uparrow$  \\
\midrule
Pooling & 20.37 & 17.23 & 0.0176 & 0.0493 & 0.1122 & 0.0804 \\ 
Residual & 20.92 & 18.62 & 0.0170 & 0.0499 & 0.1134 & 0.0810 \\
Self-Attention & 25.33 & 22.04& 0.0148 & 0.0457 & 0.1293 & 0.0961 \\ 
UniArt Attention & \textbf{28.52} & \textbf{23.77} & \textbf{0.0095} & \textbf{0.0376} & \textbf{0.1457} & \textbf{0.1176} \\
\bottomrule
\end{tabular}
}
\label{tab:ablation_aggregation}
\vspace{-0.4cm}
\end{table*}

\subsection{Open-set evaluation}
To further verify the generalization capability of UniArt, we evaluate our model on unseen object categories from the PartNet-Mobility benchmark. Specifically, we exclude categories such as Toilet, Laptop or TrashCan during training and only use them for testing. This setting poses a more challenging scenario since the model must synthesize both appearance and kinematics for categories not observed in the training set.

We show the results in Fig.~\ref{fig:base_qualitative}. We can see that despite some minor errors, UniArt successfully generates realistic and coherent articulations for the unseen categories, maintaining plausible motion patterns and detailed appearances. Despite never encountering the articulation pattern of these objects during training, the model demonstrates strong generalization by accurately synthesizing their structural parts and corresponding kinematics. This confirms UniArt’s capability to handle diverse object categories in an open-set scenario, highlighting its robustness and flexibility for practical applications.

% \begin{figure}[t]
% \centering
% \includegraphics[width= \textwidth]{fig/qualitative_ood.pdf}
% \caption{Qualitative results on unseen categories. It can be observed that the articulated objects generated by our method exhibit good consistency with the input images in both appearance and geometry, while previous retrieval-based methods fail to generate sound results.}
% \label{fig:openset_qualitative}
% \vspace{-0.5cm}
% \end{figure}

\subsection{Ablation Study}
We conduct ablation studies to analyze the contribution of different components in UniArt. We only report the joint evaluation metrics for the page limit.

\noindent \textbf{Effectiveness of Uni-encoding of Geometry and Articulation.}
UniArt aggregates information from geometry and articulation branches through a sparse structure attention to enforce joint geometry-articulation consistency. In this section, we explore an alternative aggregation strategy, vanilla aggregation, where we simply concatenate the features followed by convolution layers for dimensional alignment. 

As shown in Tab.~\ref{tab:ablation}, the vanilla concatenation approach yields a drop in generation quality (RS-$d_{\text{PSNR}}$ decreases from 28.52 to 23.75 and AS-$d_{\text{PSNR}}$ decreases from 23.77 to 21.37), indicating that simple channel stacking fails to align part-specific geometry and articulation information. By contrast, our sparse structure attention design makes a good consistency between articulation and geometry.

\noindent \textbf{Effectiveness of 3D Shape Prior.}
In our implementation, we utilize a shape prior trained from large-scale 3D generative models to make easier modeling of 3D shapes and help prevent unrealistic geometries.
In this section, we remove the pretrained 3D shape prior and train UniArt purely from scratch on PartNet-Mobility. We can see from Tab.~\ref{tab:ablation} that all of the metrics degrade significantly. This demonstrates that leveraging a large-scale 3D prior is crucial for stabilizing geometry–articulation interactions.

\noindent \textbf{Effectiveness of part-wise UniArt Attention.}
To further validate the design choice of the sparse structure attention used for fusing geometric and articulation features, we compare it against three alternative aggregation strategies: pooling, residual fusion, and vanilla self-attention. The quantitative results are summarized in Tab.~\ref{tab:ablation_aggregation}.
We observe that the naive pooling and residual fusion strategies perform poorly across all evaluation metrics, leading to a noticeable reduction in both visual fidelity and geometric accuracy. The vanilla self-attention achieves moderate improvements but still lags behind our attention design. In particular, the proposed part-wise spatial attention achieves the highest scores on all metrics, improving AS-$d_{\text{PSNR}}$ from 22.04 to 23.77 and reducing AS-$d_{\text{CD}}$ from 0.0457 to 0.0376. The OpenShape alignment score also increases by nearly 0.0215 in articulated states, indicating stronger consistency between the generated shapes and their corresponding images.
These results confirm that our sparse structure attention dynamically attends to part-level dependencies, effectively aligning geometry and motion features. This design enhances local structural coherence and globally preserves articulation plausibility, leading to substantial gains in both appearance quality and 3D consistency.

\section{Conclusion}
\label{conclusion}
In this paper, we addressed the challenge of generating articulated objects with coherent geometry, part decomposition, and functional articulation. Existing methods often rely on retrieval-based pipelines, which lead to geometry mismatches and limited category coverage. To overcome these limitations, we proposed UniArt, an end-to-end diffusion-based framework that unifies geometry generation, part segmentation, and URDF prediction into a single model. By formulating segmentation and articulation inference as open-set tasks, UniArt is capable of generalizing to unseen categories and capturing diverse part structures. Extensive experiments on PartNet-Mobility benchmarks demonstrated that our approach significantly outperforms existing baselines, both in mesh fidelity and articulation accuracy, particularly under open-set evaluation.
\clearpage
\setcounter{page}{1}
% \maketitlesupplementary

\section{Additional Dataset Details}

\subsection{PartNet-Mobility Benchmark}
We conduct all experiments on the PartNet-Mobility dataset, which provides high-quality 3D meshes with fine-grained part annotations and physically grounded articulation parameters.  
Following common practice \cite{liu2024singapo, wu2025dipo}, we select 7 common object categories, including Storage, Table, Refrigerator, Dishwasher, Oven, Washer and Microwave during training.
Each articulated object contains one or more movable parts with annotated joint types (revolute, prismatic, or fixed) and motion parameters.

\subsection{Data Preprocessing}
All meshes are normalized to fit into a unit cube with the base aligned to the canonical coordinate system.  
For each articulated object, we extract the geometry, part segmentation, and motion axes from the provided labels, and voxelize both the static (rest) and the articulated (moved) states at a resolution of $64^3$ voxels.
During preprocessing, we discard incomplete scans or objects with missing part hierarchies.
Following prior works, we randomly sample five articulation states per object by applying random joint rotations or translations within the physical limits described in the dataset annotations.

\subsection{Dataset Split}
We use dataset splits that ensure category-level separation for testing open-set generalization.  
Specifically, for \textbf{closed-set} evaluation, we follow the official PartNet-Mobility split described in Singapo \cite{liu2024singapo}. For \textbf{open-set} evaluation, we exclude certain categories (e.g., toilet, trash can) entirely from training and use them only for testing.  
This setup allows us to evaluate both generalization across unseen categories and articulation configurations.

\begin{figure*}[!t]
\centering
\includegraphics[width=0.8\textwidth]{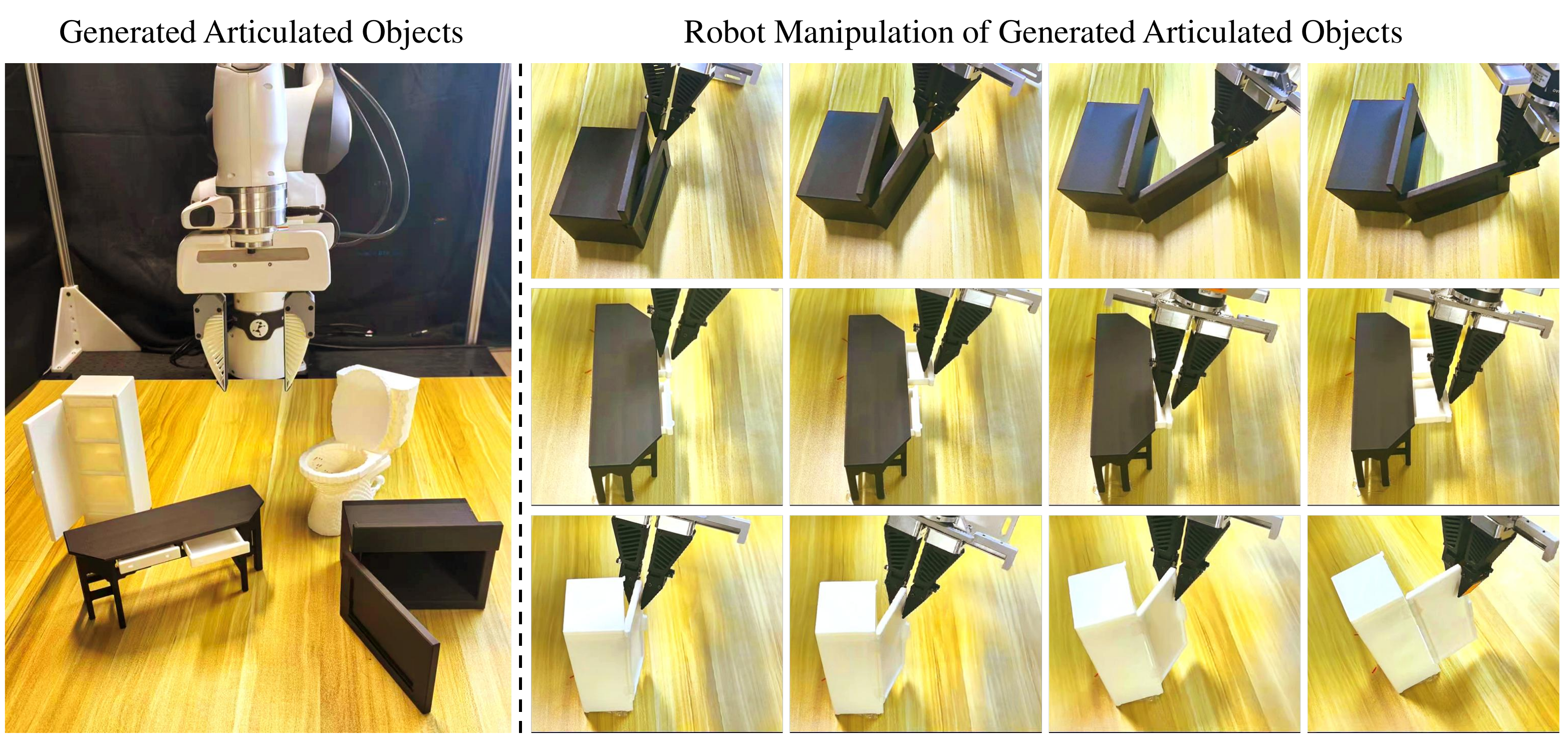}
\caption{Application in the robotic manipulation.}
\label{fig:real}
\end{figure*}

\begin{table*}[h]
\centering
\caption{\textbf{Open-Set Generalization Across Categories.}
UniArt maintains high performance on unseen categories, while existing methods experience substantial quality degradation.}
\resizebox{0.9\textwidth}{!}{
\begin{tabular}{l|cccccc}
\toprule
\multirow{2}{*}{Method} &
\multicolumn{2}{c}{Appearance-Articulation} &
\multicolumn{2}{c}{Geometry-Articulation} &
\multicolumn{2}{c}{Shape-Articulation} \\
\cmidrule(lr){2-3} \cmidrule(lr){4-5} \cmidrule(lr){6-7}
 & RS-$d_{\text{PSNR}}$ $\uparrow$ & AS-$d_{\text{PSNR}}$ $\uparrow$
 & RS-$d_{\text{CD}}$ $\downarrow$ & AS-$d_{\text{CD}}$ $\downarrow$
 & RS-$d_{\text{OpenShape}}$ $\uparrow$ & AS-$d_{\text{OpenShape}}$ $\uparrow$ \\
\midrule
Singapo~\cite{liu2024singapo} & 17.14 & 14.26 & 0.0367 & 0.0962 & 0.0989 & 0.0761 \\
\textbf{UniArt (ours)} & \textbf{22.31} & \textbf{19.05} & \textbf{0.0173} & \textbf{0.0501} & \textbf{0.1114} & \textbf{0.0813} \\

\bottomrule
\end{tabular}
}
\label{tab:supp_openset}
\end{table*}

\begin{table*}[h]
\centering
\caption{\textbf{Effect of the Multi-View Conditioning.}
Although conditioned on two or three views instead of one, improvements are marginal (2\% PSNR gain), suggesting that UniArt’s unified latent representation already captures sufficient spatial correlations from a single view.}
\resizebox{0.9\textwidth}{!}{
\begin{tabular}{l|cccccc}
\toprule
\multirow{2}{*}{Setting} &
\multicolumn{2}{c}{Appearance} &
\multicolumn{2}{c}{Geometry} &
\multicolumn{2}{c}{Shape-Image Alignment} \\
\cmidrule(lr){2-3} \cmidrule(lr){4-5} \cmidrule(lr){6-7}
 & RS-$d_{\text{PSNR}}$ $\uparrow$ & AS-$d_{\text{PSNR}}$ $\uparrow$
 & RS-$d_{\text{CD}}$ $\downarrow$ & AS-$d_{\text{CD}}$ $\downarrow$
 & RS-$d_{\text{OpenShape}}$ $\uparrow$ & AS-$d_{\text{OpenShape}}$ $\uparrow$ \\
\midrule
\textbf{UniArt} & 28.52 & 23.77 & 0.0095 & 0.0376 & \textbf{0.1457} & 0.1176 \\
\textbf{UniArt w/ multi-view} & \textbf{29.35} & \textbf{24.41} & \textbf{0.0089} & \textbf{0.0334} & 0.1456 & \textbf{0.1190} \\

\bottomrule
\end{tabular}
}
\label{tab:supp_multi_view}
\end{table*}

\section{Additional Training Details}

\subsection{Diffusion Framework Setup}
UniArt is implemented using a conditional 3D diffusion framework built upon rectified flow model \cite{lipman2022flow}. Instead of employing a stochastic diffusion process, UniArt is trained under the flow matching paradigm for generative modeling. 
Formally, a rectified flow defines a continuous and \emph{deterministic} velocity field $\mathbf{v}_\theta(\mathbf{z}_t, t \mid \mathbf{c})$ that transports samples from a simple base distribution (e.g., $\mathbf{z}_0\sim\mathcal{N}(\mathbf{0},\mathbf{I})$) toward the complex data distribution $\mathbf{z}_1\sim p_{\text{data}}(\mathbf{z}\mid\mathbf{c})$ through the integral trajectory
\begin{equation}
\frac{d\mathbf{z}_t}{dt} = \mathbf{v}_\theta(\mathbf{z}_t, t \mid \mathbf{c}), \quad t\in[0,1].
\end{equation}

Here, $\mathbf{z}_t$ represents the unified latent encoding of geometry and articulation within our 3D voxel space, and $\mathbf{c}$ denotes the conditioning input (an RGB image or multi‑view features).

Following~\cite{lipman2022flow}, the model is trained via the \emph{flow‑matching objective}:
\begin{equation}
\mathcal{L}_{\text{FM}}
  = \mathbb{E}_{t,\mathbf{z}_0,\mathbf{z}_1}
     \big[\, \| \mathbf{v}_{\theta}(\mathbf{z}_t, t \mid \mathbf{c}) -
     (\mathbf{z}_1-\mathbf{z}_0) \|_2^2 \,\big],
\end{equation}
where $\mathbf{z}_t=(1-t)\mathbf{z}_0+t\mathbf{z}_1$ is sampled along a straight path between latent pairs from the prior and the target data manifold. 
This loss encourages the predicted vector field to align with the optimal transport field that deterministically maps the Gaussian prior to the data distribution.

In contrast to diffusion models that rely on stochastic denoising with hundreds of sampling steps, flow matching learns a single deterministic ODE, significantly simplifying both training and generation. 
During inference, we integrate the learned velocity field from $t=0$ to $t=1$ using an adaptive‑step Euler or Heun solver to generate articulated 3D objects in the latent voxel space.
Empirically, this rectified‑flow formulation yields reduction in sampling cost while improving stability and coherence between geometry and motion.

\subsection{Optimization and Hyperparameters}
We use the AdamW optimizer with an initial learning rate of $1\times10^{-4}$, $\beta_1=0.9$, $\beta_2=0.999$, and a weight decay of 0.01.  
Training is performed for 800K iterations with a batch size of 8 per GPU across 8 NVIDIA A100 GPUs. 
We adopt exponential moving average (EMA) weights with a decay of 0.9995 for stable evaluation.  

The 3D shape prior is obtained from a pretrained TRELLIS model \cite{xiang2025structured}, which is finetuned during UniArt training and used to guide geometric consistency.

The final training objective $\mathcal{L}_{\text{total}}$ combines multiple components:
\begin{equation}
    \mathcal{L}_{vae} = \mathcal{L}_{geo} + \lambda_1\mathcal{L}_{art} + \lambda_2\mathcal{L}_{kl}.
\end{equation}
where $\mathcal{L}_{geo}$ represents voxel-based reconstruction loss between predicted and GT occupancy grids, $\mathcal{L}_{art}$ represents articulation parameter regression loss (joint axis, limits), and $\mathcal{L}_{kl}$ represents structure alignment loss enforcing consistency between geometry and articulation embeddings.  
Empirically, we set $(\lambda_1, \lambda_2) = (1.0, 0.5)$.

\section{Additional Experiments}

\subsection{Open-Set Generalization Across Categories}
To further examine UniArt’s open-set generalization ability, we evaluate on categories unseen during training (toilet, trash can, etc). We only choose the categories with more than 10 objects.
For fair comparison, we adapt Singapo to an open-set paradigm. Firstly, we remove its category-specific modules when encoding the articulation parameters. Specifically, Singapo could only generate parts with specific categories like handles. We remove this encoding for its articulation encoding and retrain it using only the articulation parameters shared across all object categories and part categories, thus allowing it to infer on unseen object types.
Results in ~\ref{tab:supp_openset} show that UniArt maintains over 85\% of its closed-set performance across all metrics, while SINGAPO drop to below 60\%, confirming that defining articulation as an open-set problem significantly enhances category transferability.

\subsection{Multi-View Conditioning}
We additionally train a variant of UniArt with multi-view image conditioning to test its scalability.  As shown in Table.~\ref{tab:supp_multi_view}, although conditioned on two or three views instead of one, improvements are marginal (2\% PSNR gain), suggesting that UniArt’s unified latent representation already captures sufficient spatial correlations from a single view.

\subsection{Real-World Deployment in Robot Simulation}
To validate the practical value of UniArt beyond offline metrics, we deploy the generated articulated assets in both a physics simulator and a real-world robotic manipulation setup. Specifically, in the simulator, we export each mesh together with its predicted URDF directly into MuJoCo and PyBullet, where they are instantiated without any manual post-processing; a scripted impedance controller then executes three canonical primitives, hinge opening, slider pulling, and compound flip-and-rotate motions, while success is recorded when the commanded joint approaches at least 70\% of its predicted range without self-collision. In the real-robot experiments, we use a 3D Printer to print the generated part and assemble it according to the URDF file. Then, we use an open-source articulation manipulation policy to open the generated objects. The results shown in Fig.~\ref{fig:real} prove the effectiveness of our method.

{
    \small
    \bibliographystyle{ieeenat_fullname}
    \bibliography{main}
}

% WARNING: do not forget to delete the supplementary pages from your submission 
% \input{sec/X_suppl}

\end{document}